\documentclass[fleqn,10pt]{wlscirep}
\usepackage[utf8]{inputenc}
\usepackage[T1]{fontenc}
\usepackage{amsthm}  % Add this package

\newtheorem{definition}{Definition}  % Define the environment

\title{Learning-based agricultural management in partially observable environments subject to climate variability}

\author[1]{Zhaoan Wang}
\author[1,*]{Shaoping Xiao}
\author[1]{Junchao Li}
\author[2]{Jun Wang}

\affil[1]{Department of Mechanical Engineering, Iowa Technology Institute, University of Iowa,
            3131 Seamans Center,
            Iowa City,
            52242, 
            Iowa,
            USA}

\affil[2]{Department of Chemical and Biochemical Engineering, Iowa Technology Institute, University of Iowa,
            4133 Seamans Center, 
            Iowa City,
            52242, 
            Iowa,
            USA}

\affil[*]{shaoping-xiao@uiowa.edu}


\begin{abstract}
Agricultural management, with a particular focus on fertilization strategies, holds a central role in shaping crop yield, economic profitability, and environmental sustainability. While conventional guidelines offer valuable insights, their efficacy diminishes when confronted with extreme weather conditions, such as heatwaves and droughts. In this study, we introduce an innovative framework that integrates Deep Reinforcement Learning (DRL) with Recurrent Neural Networks (RNNs). Leveraging the Gym-DSSAT simulator, we train an intelligent agent to master optimal nitrogen fertilization management. Through a series of simulation experiments conducted on corn crops in Iowa, we compare Partially Observable Markov Decision Process (POMDP) models with Markov Decision Process (MDP) models. Our research underscores the advantages of utilizing sequential observations in developing more efficient nitrogen input policies. Additionally, we explore the impact of climate variability, particularly during extreme weather events, on agricultural outcomes and management. Our findings demonstrate the adaptability of fertilization policies to varying climate conditions. Notably, a fixed policy exhibits resilience in the face of minor climate fluctuations, leading to commendable corn yields, cost-effectiveness, and environmental conservation. However, our study illuminates the need for agent retraining to acquire new optimal policies under extreme weather events. This research charts a promising course toward adaptable fertilization strategies that can seamlessly align with dynamic climate scenarios, ultimately contributing to the optimization of crop management practices.
\newline
\end{abstract}

\begin{document}

\flushbottom
\maketitle
\thispagestyle{empty}

\section{Introduction}

According to a 2022 report from the United States Department of Agriculture (USDA) \cite{Wang2022}, total farm production nearly tripled from 1948 to 2017. However, despite the growth, there remains a global food shortage. The Food and Agriculture Organization (FAO) estimated that approximately 828 million people were experiencing hunger in 2022. Given this pressing issue, it becomes imperative to leverage new technologies to boost farm production, and one such solution is Precision Agriculture (PA) \cite{Zhang2022}. Precision agriculture, also known as ``precision farming'' or ``prescription farming,'' utilizes information and technology-based agricultural management systems. These systems enable farmers to precisely tailor their soil and crop management practices to various weather/soil conditions on individual farmlands. 

In a modern community, PA is an emerging field aimed at enhancing the efficiency and sustainability of agricultural practices \cite{Duru2015}. Precision agriculture often employs advanced technologies such as remote sensing, robotics, Machine Learning (ML), and Artificial Intelligence (AI) techniques. Monitoring plant health and detecting diseases are vital aspects of sustainable agriculture. Yet, manual disease detection is labor-intensive, necessitating significant expertise, effort, and extended processing time. Researchers have turned to image recognition algorithms as a solution, achieving promising results in plant disease identification \cite{Khirade2015}. Additionally, AI's potential in forecasting crop yields has gained significant attention. Some researchers have used satellite imagery to develop models that predict yields, often incorporating crop identification maps and meteorological data. These models have been applied to forecast yields for crops such as wheat, rice, cotton, and sugarcane, especially in regions like the Indus Basin in Pakistan, demonstrating satisfactory performance \cite{Bastiaanssen2003}.

As one of the important components in PA, learning-based agricultural management represents a substantial departure from traditional farming methods, which often rely on human intuition and experience. Learning-based agricultural management adopts a more data-driven approach \cite{Kirtan2019} with the overarching goals of increasing efficiency, reducing waste, protecting the environment, and improving the sustainability of farming practices. A notable example is seen in the work of Vij and co-authors \cite{Vij2020}, who predicted the irrigation needs of farmland. They achieved this by using intelligent systems to monitor ground parameters, including soil moisture, soil temperature, and environmental conditions such as air temperature, ultraviolet rays, light radiation, and the relative humidity of the fields. Additionally, they incorporated weather forecast data sourced from the internet.

In previous studies of agricultural management, researchers traditionally collected and analyzed historical data to identify empirical regularities, which could then inform future agricultural policies and practices \cite{Kenneth2002}. With the continuous advancement of computer simulation technology, specialized software tools such as Decision Support System for Agrotechnology Transfer (DSSAT) \cite{Jones2003}, Agricultural Production Systems Simulator (APSIM) \cite{Keating2003}, and AquaCrop \cite{Steduto2009} have been developed. These simulation tools are designed to model various aspects of crop growth, yield, water, and nutrient requirements in response to environmental conditions. Particularly, DSSAT has gained widespread recognition and has been employed for over 30 years to simulate crop behavior and responses to environmental variables, making it a valuable resource for crop simulation studies. 

The above-mentioned software tools can effectively approximate the growth process of crops and predict the final yields by considering management parameters and environmental conditions such as temperature, humidity, soil properties, and other influential factors. Among those factors, nitrogen fertilizer stands out as a crucial element that can be managed and controlled. Nitrogen is the primary nutrient that profoundly affects crop growth and yield production. However, an excessive application of nitrogen fertilizer can lead to substantial detrimental effects on the environment \cite{Mark2011}, including nitrate leaching. Therefore, it becomes imperative to implement effective nitrogen management strategies to balance optimizing crop yields, minimizing environmental damage, and sustaining farmers' income.

As one subset of ML, Reinforcement Learning (RL) empowers computer programs, acting as agents, to control unknown and uncertain dynamical systems while pursuing specific tasks \cite{Li2023, Cai2023}. This approach has garnered increasing attention from researchers interested in determining optimal strategies for agricultural management \cite {Overweg2021}. Notably, the DSSAT, a widely recognized agricultural simulation tool \cite{Jones2003}, has been extended to a realistic simulation environment known as Gym-DSSAT \cite{Romain2022}. In this environment, RL agents can effectively learn fertilization and irrigation management strategies when provided with soil property data and weather history/forecast information. Specifically, Gautron \textit{et al.} \cite{Romain2022} proved that RL agents could discover interesting crop management policies in simulated conditions and gym-DSSAT and simulated worldwide growing conditions. In addition, Wu \textit{et al.} \cite{Wu2022} recently demonstrated that the RL-trained policies outperformed empirical methods, resulting in higher or similar crop yields while using fewer fertilizers. Sun \textit{et al.} \cite{Sun2017} conducted research wherein they formulated a reinforcement learning-driven irrigation control method. This technique has the potential to substantially enhance net gains by accounting for both crop yield and water expenditure.

The aforementioned works \cite{Romain2022, Wu2022, Sun2017} predominantly assumed the agricultural environment was fully observable, leading to the mathematical formulation of corresponding RL problems as Markov Decision Process (MDP) problems. In MDP, each state of the environment is expected to encompass all the necessary information for the agent to determine the best action for optimizing the objective function. However, questions arise regarding whether the state variable listed in Gym-DSSAT can comprehensively represent the state of the agricultural environment \cite{Puterman2014}. Moreover, certain state variables, such as the index of plant water stress, daily nitrogen denitrification, and daily nitrogen plant population uptake, may be challenging to measure and access. 

This issue mirrors many real-world applications where agents lack complete knowledge to determine the environment's state precisely. In such cases, agents often only have access to uncertain or incomplete observations of the states. This challenge may be addressed by the Partially Observable Markov Decision Process (POMDP) framework \cite{Astrom1965}. While POMDP was mentioned in the context of Gym-DSSAT \cite{Romain2022} during its introduction, the specific solution was not detailed. Recently, Tao \textit{et al.} \cite{Tao2022} employed Imitation Learning (IL) to develop management policies that require only a minimal number of state variables by mirroring the actions of the RL policies learned with full observation. They discovered that the policies,  after being learned under partial observation, demonstrated decisions almost identical to those trained with RL under full observation. 

On the other hand, climate variability is another critical factor in agriculture and its management, encompassing changes in temperature, precipitation, wind patterns, and other meteorological elements occurring over various temporal and spatial scales \cite{IPCC2014}.  Weather conditions, especially extreme weather events, can significantly impact final crop yields. For example, Motha and Baier \cite{Motha2005} conducted a study analyzing the time series of corn yields from 1895 to 2002 in the state of Iowa. They identified substantial agricultural losses in 1988 due to one of the worst droughts during the growing season in modern history. Additionally, flooding caused almost a 50\% drop in Iowa's corn production in 1993 compared to the previous year. Therefore, it is crucial for the learning agent of agricultural management to be adaptive to climate variability.   

This paper presents a framework for optimizing nitrogen fertilization while considering the agricultural environment as partially observable. Additionally, we investigate the impact of climate variability on nitrogen fertilization and crop production. Our contributions are twofold. 

First, our study demonstrates the effectiveness of formulating the agricultural environment as a POMDP in generating superior policies (i.e., management strategies) compared to using an MDP, which has been the assumption in most prior works \cite{Romain2022, Wu2022, Sun2017}. This conclusion contrasts with the findings of Tao \textit{et al.} \cite{Tao2022}, where the agent was initially pre-trained in MDP and subsequently in POMDP through imitation learning, resulting in similar policy performance. Furthermore, our approach enables the agent to learn optimal policies within the POMDP framework directly. Specifically, we employ a model-free RL method that incorporates RNNs to solve POMDP problems. 

Secondly, we investigate and quantify the influence of climate variability on agricultural practices and crop production. We particularly emphasize two extreme events: a heatwave in 1983 and a drought in 1988. These case studies illustrate the adaptability of RL agents to learn optimal nitrogen fertilization policies under extreme conditions. To the best of our knowledge, no similar systematic investigations have been reported in the literature on learning-based agricultural management.

The structure of this paper is organized as follows. In Section 2, we present the formulations of MDP and POMDP and introduce Q learning, a model-free RL method. Section 3 sets up and compares various MDP and POMDP models. In Section 4, we delve into the impact of climate variability on crop yield and nitrogen fertilizer usage, including the study of two extreme weather events. Finally, we conclude the paper in Section 5 and outline avenues for future research.

\section{Methodology}

\begin{figure}
\centering
\resizebox*{7cm}{!}{\includegraphics{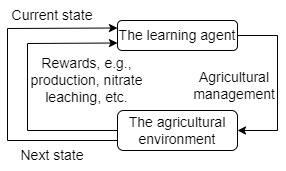}}
\caption{The reinforcement learning approach. } \label{fig:RL}
\end{figure}

In this study, we utilize an RL approach. During a learning process, as illustrated in Figure~\ref{fig:RL}, the agent interacts with the agricultural environment by taking actions in agricultural management and receiving rewards as feedback. The MDP, a mathematical framework that describes the environment and its interaction with the agent, assumes the environment is fully observable. Under this assumption, the agent can completely identify the environment's current state (i.e., configuration) and learn how to make optimal decisions accordingly. However, in most real-world applications, the agent only receives incomplete information, which cannot be used to identify the current state of the environment. Therefore, the MDP is unsuitable, and the POMDP must be adopted. This section will provide mathematical definitions for both MDP and POMDP. Subsequently, we will introduce a model-free RL method called Q learning and then extend it by incorporating RNNs to solve POMDP problems.

\subsection{MDP and POMDP}

\begin{definition}[MDP]\label{def:MDP} 
An MDP can be generally denoted by a tuple $\mathcal{P}=\left(S, A, T,s_0, R \right)$, where:

\begin{itemize}
\item $S = \{s_1, . . . , s_n\} $ is a finite set of states. 
\item $A = \{a_1, . . . , a_m\}$ is a finite set of actions. In particular, $A(s)$ represents a set of actions that the learning agent can take at state $s$.
\item $T : S\times A\times S\rightarrow \left[0,1\right]$  is a function representing the transition probability from state $s \in S$ to state $s' \in S$ after the agent takes action $a \in A(s)$. It satisfies $\sum_{s'\in S} T(s,a,s') = 1$. 
\item $s_0 \in S$ is the initial state.
\item $R : S \times A \times S \rightarrow R$ is a reward function as $R(s, a,s')$.  The reward function may have other formulations, such as $R(s')$ or $R(s,a)$.
\end{itemize}
\end{definition}

When we use RL approaches to solving MDP problems, it is crucial to grasp how an agent interacts with its environment. When the agent engages with the environment, it makes decisions based on its knowledge of the current state, denoted as $s$. Once the agent selects an action, represented as $a$, the environment transitions to a new state, which we denote as $s'$. This transition occurs with a probability determined by the function $T(s,a,s')$. Simultaneously, the agent receives immediate feedback in the form of a reward, denoted as $R(s,a,s')$.

\begin{definition}[POMDP]\label{def:POMDP} 
A POMDP can be generally denoted by a tuple $\mathcal{P}=\left(S, A, T,s_0, R, O, \Omega \right)$, where $S, A, T, S_0,$ and $R$ are defined as the same in MDP (Definition~\ref{def:MDP}), and 

\begin{itemize}
\item $O = \{o_1, . . . , o_z\}$ is a finite set of observations. $O(s)$ is a set of possible observations the agent can perceive at state $s$.
\item $\Omega : S \times A \times O \rightarrow \left[0,1\right]$ is a function representing the observation probability that the agent can perceive at state $s' \in S$ after taking action $a \in A(s)$. This function satisfies $\sum_{o\in O} \Omega(s',a,o) = 1$.
\end{itemize}
\end{definition}

When the agent receives only partial information about the environment's current state, its decision-making relies on both past and current observations. Once the agent takes a selected action and transitions to the next state $s'$, it perceives a new observation $o \in O(s')$ with a probability described by $\Omega(s', a, o)$. The primary objective of an intelligent agent is to learn an optimal policy that maximizes the expected return, as formulated below. This expected return represents the cumulative rewards starting from the current state.

\begin{equation}\label{eq:expReturn} 
U(s) = \mathbb{E} \left[\sum_{t=0}^{\infty} \gamma^t R(s_t, a_t, s_{t+1}) \Big \vert s_{t=0} = s\right]
\end{equation}
where $s_{t}$ denotes the agent’s state at time $t$. $\gamma \in [0,1]$ is the discount factor to balance the importance between immediate and future rewards. 

There have been several model-based approaches \cite{Li2023} to solving POMDP problems. These approaches commonly seek optimal policies in the belief state space rather than the state space defined in the POMDP (Definition~\ref{def:POMDP}). A belief state is a probability distribution encompassing all the possible states where the agent could be. It can be dynamically updated based on transition and observation probabilities during the learning process. When using the model-based approach, a POMDP problem transforms into a search for optimal policies within a corresponding MDP defined in the belief state space. However, in this study, the agent lacks knowledge of the transition and observation probabilities, rendering model-free RL methods an appropriate choice.

\subsection{Q-learning and Deep Q-Network}

The expected return in Equation (\ref{eq:expReturn}) also defines the state value function, denoted as $V(s)$, at state $s$. Similarly, there is another value function, $Q(s,a)$, referred to as state-action value, action value, or $Q$ value. It represents the total reward an agent can accumulate over the long run after taking action $a$ at state $s$. In the realm of RL, there are two main categories of methods: value-based and policy-based. Policy-based methods seek optimal policies directly, while value-based RL methods focus on determining optimal value functions. Subsequently, optimal policies can be derived through greedy action selection.

Q-learning \cite{Watkins1992} is a model-free, value-based RL method in which the agent tends to achieve optimal state-action values. As a tabular method, the na\"ive Q-learning employs a Q-table to store $Q$ values and quantify the best action with the highest $Q$ value for the agent to choose. On the other hand, $Q$ values in the Q-table are updated via bootstrapping when the agent interacts with its environment. In each episode, $Q(s,a)$ is updated at each step as below after taking action $a$ at state $s$, following the Bellman equation \cite{Sutton2018}.
\begin{equation}\label{eq:QlearningValueFunc}
Q_{new}(s,a)=Q(s,a) + \alpha [R(s,a,s') + \gamma \max_{a'\in A}Q(s',a') - Q(s,a)]
\end{equation}
where $\alpha$ is the learning rate, enhancing the efficiency and stability of Q-value convergence.  

Usually, the $\epsilon$-greedy technique is employed. This means that there is a probability of $\epsilon$ for the agent to choose a non-optimal action, allowing it to explore the state space, in addition to exploiting the current policy. Once the optimal value function, $Q^*(s,a)$, is reached, the optimal policy can be extracted as $\xi^*(s) = \arg\max_{a \in A} Q^*(s,a)$.

However, tabular Q-learning becomes unsuitable when dealing with a large or infinite state space, such as agricultural environments. This challenge can be addressed by replacing the Q-table with deep neural networks (DNNs), known as Q-networks, to estimate $Q$ values. This approach falls under the umbrella of Deep Reinforcement Learning (DRL) \cite{Arulkumaran2017}, and in this study, we employed the Deep Q-Network (DQN) \cite{Mnih2013}, which is an extension of Q-learning.

Deep Q-Network consists of two Q-networks: an evaluation Q-network, denoted as $Q_e(s, a;\theta_e)$, and a target Q-network, denoted as $Q_t(s,a; \theta_t)$. Here, $\theta_e$ and $\theta_t$ represent the network weights, which are to be trained and updated through the experience replay memory \cite{Lin1992}. During the learning process, an experience is generated at each step in the form of $(s, a, s', R)$ and stored in a memory pool. Simultaneously, a set of these experiences, referred to as a mini-batch, is selected from the memory pool to train and update the evaluation Q-network. The Bellman equation presented in Equation~(\ref{eq:QlearningValueFunc}) is modified as follows. 
\begin{equation} \label{eq:Deep Q-learning}
Q_{new}(s,a) = Q_e(s,a; \theta_e)+\alpha \left[ R(s, a, s')+\gamma \max_{a' \in A} Q_t(s',a'; \theta_t)-Q_e(s,a; \theta_e) \right]
\end{equation}

It is worth mentioning that the target Q-network is not trained by holding fixed weights until copying from the evaluation Q-network, i.e., $\theta_t = \theta_e$, once in a while. Instead, this approach allows the target network to update incrementally, known as a soft update. In a soft update, the target network slowly tracks changes in the evaluation network, which helps improve stability and convergence during training.  

In MDP, the agent has complete knowledge of the environment, and the value function assesses available actions at the current state for decision-making. However, when the environment becomes partially observable, the agent must rely on a history of perceived information, typically a sequence of observations, to make informed decisions. As a result, the Q-networks in DQN take this sequence of observations as the input and produce corresponding $Q$ values. Furthermore, the policy or agent function now maps a sequence of observations to an action. In such cases, RNNs emerge as strong candidates for Q-networks because of their proficiency in handling sequential or time-series data.  

In a recent study by Li \textit{et al.} \cite{Li2023}, an RNN-based DQN was proposed for robotics motion planning in partially observable environments. They utilized Long Short-Term Memory (LSTM) \cite{Hochreiter1997} in their Q-networks to process sequences of observations. In this study, we opt for Gated Recurrent Units (GRU) \cite{Cho2014}, another advanced RNN architecture, within our Q-networks to model temporal dependencies among observations, incorporating feedback loops within the network structure. We conduct a comparison between LSTM-based DQN and GRU-based DQN, and both approaches yield similar results. However, due to its forget gate and the absence of an output gate, a GRU cell has fewer parameters, which results in increased efficiency and reduced training time compared to LSTM. 

\begin{figure}
\centering
\resizebox*{9cm}{!}{\includegraphics{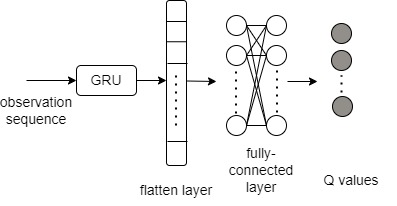}}
\caption{GRU-based Q-network architecture } \label{fig:QNet}
\end{figure}

The architecture of the GRU-based Q-network is depicted in Figure~\ref{fig:QNet}. The sequence of observations, denoted as  $\mathbf{o}_{t} = (o_{t-j}, o_{t-j+1}, ..., o_{t})$, has a length of $j+1$. Since Q networks take the observation sequences as input, we redefine the evaluation Q-network as $Q_E(\mathbf{o}_t, a_t; \theta_E)$ and the target Q-network as $Q_T(\mathbf{o}_t, a_t; \theta_T)$. During the learning process, after the agent reaches the next state, it receives a reward $R_t$ and perceives a new observation $o_{t+1}$. A new observation sequence is then generated as $\mathbf{o}_{t+1} = (o_{t-j+1}, o_{t-j+2}, ..., o_{t+1})$. Consequently, a new data sample (or experience), e.g., $(\mathbf{o}_{t}, a_t, R_t, \mathbf{o}_{t+1})$, is formed and recorded in the replay memory to update the evaluation Q-network. Furthermore, Equation~(\ref{eq:Deep Q-learning}) can be expressed as 
\begin{equation}\label{eq:DRQN Q value}
Q_{new}(\mathbf{o}_t, a_t) = Q_E(\mathbf{o}_t,a_t; \theta_E)+\alpha \left[ R_t+\gamma \max_{a_{t+1}} Q_T(\mathbf{o}_{t+1},a_{t+1}; \theta_T)-Q_E(\mathbf{o}_t,a_t; \theta_E) \right]
\end{equation}

After the learning finishes, the evaluation Q-network is converged, and the optimal Q values can be estimated. Furthermore, the optimal policy can be derived by
\begin{equation}\label{eq:POMDP policy}
\xi^*(\mathbf{o}) = \arg\max_{a \in A} Q^*(\mathbf{o},a).
\end{equation}

\section{Agriculture management as a POMDP problem}

In this study, we utilize maize crop growth in Iowa as a case study to demonstrate that the agricultural environment, represented by DSSAT, is partially observable. Our study encompasses the years 1965, 1980, 1999, and 2020. We obtain the corresponding weather data from the Iowa State University Soil Moisture Network \cite{ISUSMN}, which includes daily maximum temperature, minimum temperature, solar radiation, and precipitation. However, due to a lack of comprehensive data, we rely on the soil property data from 1999 and apply it to the other years in our study. The basic Gym-DSSAT input file from 1999 weather and soil data and the DRL code for this research can be found in our GitHub repository. (https://github.com/ZhaoanWang/Learning-based-agricultural-Management).

\subsection{Model setup}

The Gym-DSSAT employs factored representations, utilizing a total of 28 internal variables, as detailed in Table~\ref{table:stateVar}. Many studies \cite{Romain2022, Wu2022} have used these variables to represent the agricultural environment's state. They have framed learning-based agricultural management problems as MDP problems, assuming full observability of the environment. This assumption implies that the environment possesses the Markov property, allowing the agent to make decisions based on the immediately-received state variables. 
However, it is important to note that there is no conclusive evidence to demonstrate that these 28 internal variables can entirely determine the state of the agricultural environment. Furthermore, not all of these variables are easily observable or accessible. This study uses the first 10 variables from Table~\ref{table:stateVar} as observation variables. This approach enables the agent to make decisions based on both current and previous observations. Consequently, the original problem is transformed into a POMDP problem.

\begin{table}
\centering
%\begin{tabular} {||c c||} 
\begin{tabular} {| p{2.5cm} | p{10cm} | }
 \hline
 Variable & Description \\ [0.5ex] 
 \hline\hline

 \hline
 \textbf{cumsumfert} & cumulative nitrogen fertilizer applications (kg/ha) 
 \\ \hline
 \textbf{dap} & days after planting  \\
 \hline
 \textbf{istage} & DSSAT maize growing stage  \\ 
 \hline
  \textbf{pltpop} & plant population density (plant/m$^2$) \\ 
 \hline
 \textbf{rain} & rainfall for the current day (mm/d)	 \\
 \hline
 \textbf{sw} & volumetric soil water content in soil layers (cm$^3$ [water] / cm$^3$ [soil]) \\
 \hline
 \textbf{tmax} & maximum temperature for the current day ($^{\circ}$C)  \\
 \hline
 \textbf{tmin} & minimum temperature for the current day ($^{\circ}$C) \\
 \hline
 \textbf{vstage} & vegetative growth stage (number of leaves)   \\
 \hline
 \textbf{xlai} & plant population leaf area index   \\
 \hline
 cleach & cumulative nitrate leaching (kg/ha)   \\
 \hline
 cnox & cumulative nitrogen denitrification (kg/ha)   \\
 \hline
 dtt & growing degree days for the current day (C/d)   \\
 \hline
 es & actual soil evaporation rate (mm/d)   \\
 \hline
 grnwt & grain weight dry matter (kg/ha)   \\
 \hline
 nstres & index of plant nitrogen stress    \\
 \hline
 pcngrn & massic fraction of nitrogen in grains    \\
 \hline
 rtdep & root depth (cm)   \\
 \hline
 runoff & calculated runoff (mm/d)   \\
 \hline
 srad & solar radiation during the current day (MJ/m$^2$/d)   \\
 \hline
 swfac & index of plant water stress   \\
 \hline
 tleachd & daily nitrate leaching (kg/ha)   \\
 \hline
 tnoxd & daily nitrogen denitrification (kg/ha)   \\
 \hline
 topwt & above the ground population biomass (kg/ha)   \\
 \hline
 totir & total irrigated water (mm)   \\
 \hline
 trun & daily nitrogen plant population uptake (kg/ha)   \\
 \hline
 wtdep & depth to water table (cm)   \\
 \hline
 wtnup & cumulative plant population  nitrogen uptake (kg/ha)  \\
 \hline 
\end{tabular}

\caption{Internal state variables of the agricultural environment. }\label{table:stateVar}
\end{table}

This section explores four problems using different MDP and POMDP models. We then compare corn yields and N fertilizer usages resulting from the optimal policies ${-}$ i.e., management strategies learned by the agent ${-}$ in the previously mentioned years. These problem types are as follows:
\begin{itemize}
\item MDP-28: Markov decision process problems with all 28 internal variables as state variables. 
\item POMDP-28: Partially observable Markov decision process problems with all 28 internal variables as observation variables.
\item MDP-10: Markov decision process problems with the first 10 internal variables as state variables.
\item POMDP-10: Partially observable Markov decision process problems with the first 10 internal variables as observation variables.
\end{itemize}

Given that maize crops in Iowa are typically rain-fed \cite{Wu2022}, this study excludes daily irrigation considerations and concentrates on nitrogen fertilization. Consequently, the action space encompasses various quantities of nitrogen that can be applied in a single day. Mathematically, the action space is discretized as ${10k (kg/ha)}$ nitrogen input, where $k$ ranges from 1 to 20. 

In a given day `$d_t$,' after taking action, which involves applying an amount of nitrogen $N_t$, the agent receives a reward defined as:
\begin{equation} \label{eq:AgReward}
R(d_t, N_t)= \left\{ \begin{array}{cc} w_1 Y - w_2 N_t - w_3 L_t & \mbox{at harvest}
\\ -w_2 N_t - w_3 L_t & \mbox{otherwise} \end{array} \right.
\end{equation}
where $Y$ represents the corn yield at harvest, and $L_t$ denotes nitrate leaching on a particular day $t$. The weight coefficients, $w_1$ and $w_2$, are determined by the prevailing prices of corn and nitrogen input in each simulated year, as listed in Table~\ref{table:RewardCoe}. Particularly, $w_1$ corresponds to the price of corn per kilogram \cite{USDA}, and $w_2$ is based on the price of nitrogen per kilogram, calculated using 45\% urea nitrogen to determine the price of 100\% nitrogen \cite{ISU}. In addition, $w_3$ is the weight assigned to nitrate leaching, calculated as a multiple of $w_2$. The specific multiple is 5, as indicated in \cite{Wu2022}.

\begin{table}
\centering
%\begin{tabular}{||c c c c||} 
\begin{tabular} {| p{2cm} | p{2cm} | p{2cm} | p{2cm} | }

 \hline
 Year & w1 & w2 & w3  \\ [0.5ex] 
 \hline\hline

 \hline
 1965 & 0.03819  & 0.26   & 1.04
 \\ 
 \hline
 1980 & 0.07953 & 0.49   & 1.96 
 \\
 \hline
 1999 & 0.07087  & 0.39   & 1.95   
 \\
 \hline
 2020 & 0.1827 & 0.87  & 3.48
 \\
 \hline 
\end{tabular}

\caption{Weight coefficients used in reward functions}\label{table:RewardCoe}
\end{table}

When using DQN to solve MDP problems, as defined above, Q-networks are fully connected networks that take state variables as the input and output $Q$ values for each action. The network architecture consists of 3 hidden layers with 256 units in each layer, and the rectified linear activation function (ReLU) is used. On the other hand, when solving POMDP problems, Q-networks take a sequence of observations as input, and each observation is a vector of observation variables. We test various sequence lengths and find that 5 time steps (i.e., 5 days) are the proper length. The GRU layer in the Q-networks has one hidden layer with 64 units, and its output is passed to a fully connected network, which is the same as the one used in solving MDP problems to calculate $Q$ values.

The training process includes 6000 episodes, each lasting 180 steps (i.e., days). We set the discount factor to 0.99. To update the neural networks, we utilize Pytorch and Adam optimizer \cite{Kingma2014} with an initial learning rate of 1e-5 and a batch size of 640. The simulations are conducted on a machine equipped with an Intel Core i7-12700K processor, an NVIDIA GeForce RTX 3070 Ti graphics card, and 64 GB of RAM.

\subsection{Results and discussions}

Table~\ref{table:MdpPomdpCompare} displays the accumulated rewards each year, reflecting the performance of optimal policies learned by the agent within various models. Notably, the MDP-28 model surpasses the MDP-10 model, indicating that the inclusion of 28 state variables enhances decision-making by providing more information if the agent can only access the current agricultural state (i.e., configuration). On the other hand, all POMDP models outperform the MDP-28 model as hypothesized, shedding light on the fact that the agricultural environment is only partially observable through the internal state variables listed in Table~\ref{table:stateVar}. This finding suggests that the agent benefits from leveraging a history of observations to formulate better policies. Intriguingly, the optimal policies derived from POMDP-28 and POMDP-10 exhibit striking similarity, implying that employing 10 observation variables is well-informed for the agent's decision-making in the Gym-DSSAT environment.

\begin{table}
\centering
%\begin{tabular}{||c c c c c ||} 
\begin{tabular} {| p{2cm} | p{2cm} | p{2.5cm} | p{2cm} | p{2.5cm} |
}
 \hline
 Year & MDP-28 & POMDP-28& MDP-10 & POMDP-10  \\ [0.5ex] 
 \hline\hline

 \hline
 1965 & 235  &350 & 187  & 350  
 \\ 
 \hline
 1980 & 594 &612 & 460  & 612  
 \\
 \hline
 1999 & 515 & 584 & 435 & 584 
 \\
 \hline
 2020 & 1435 & 1471 & 1200 & 1466 
 \\
 \hline 
\end{tabular}

\caption{Accumulative rewards from optimal policies trained by different models.}\label{table:MdpPomdpCompare}
\end{table}

In Table~\ref{table:1999outcomes}, we compare the outcomes of 1999, which include corn yield, nitrogen input, and nitrate leaching, obtained from the optimal policies learned by the agent within different models. Surprisingly, all four optimal policies achieve similar corn yields. However, notable differences emerge in nitrogen usage and nitrate leaching. Policies learned within POMDP models recommend significantly less nitrogen usage than those from MDP models. This reduction in nitrogen input also results in less nitrate leaching, contributing to higher accumulated rewards for the POMDP-induced policies than their MDP counterparts. 

\begin{table}
\centering
%\begin{tabular}{||c c c c||} 
\begin{tabular} {| p{3cm} | p{2cm} | p{3cm} | p{3cm} |
} \hline
 Policy from  & Yield (kg/ha) & Nitrogen input (kg/ha) & Nitrate leaching (kg/ha)\\ [0.5ex] 
 \hline
 \hline
 MDP-28  & 9247  & 360 & 0.14 
 \\ 
 \hline
 POMDP-28 & 9243  & 180 & 0.12
 \\
 \hline
 MDP-10   & 9226 & 560 & 0.20
 \\
 \hline
 POMDP-10  & 9243 & 180 & 0.12
 \\
 \hline 
 Expert policy 1 & 6236 & 56 & 0.12
 \\
 \hline 
  Expert policy 2 & 9247 & 224 & 0.26
 \\
 \hline 
\end{tabular}

\caption{Outcomes of 1999 from various optimal policies, compared to the expert policies, which respectively result in total rewards of 425 and 567.}\label{table:1999outcomes}
\end{table}

Additionally, we provide the results from two expert policies offered by Gym-DSSAT in Table~\ref{table:1999outcomes}. The first expert policy employs minimal nitrogen input and consequently yields much less corn than other policies. In contrast, the second expert policy utilizes a more substantial amount of nitrogen and yields more corn than the first expert policy, aligning closely with the outcomes of the learned optimal policies. However, this second expert policy also leads to the highest level of nitrate leaching, although its nitrogen usage is less than the optimal policies from MDP models.

\begin{figure}
\centering
\resizebox*{10cm}{!}{\includegraphics{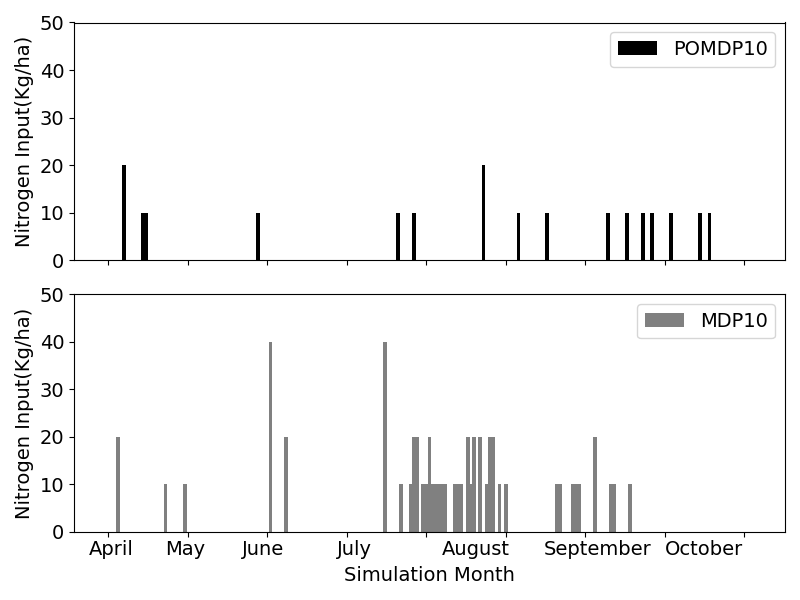}}
\caption{Comparison of fertilizer managements based on different optimal policies from MDP-10 and POMDP-10 models.}
\label{fig:1999Fertilization}
\end{figure}

As seen in Figure~\ref{fig:1999Fertilization}, we can observe the nitrogen fertilization schedules and quantities based on the optimal policies derived from MDP and POMDP models. Notably, the MDP-10 policy applies nitrogen more frequently and in larger amounts than the policy generated by the POMDP-10 model. This discrepancy likely arises from the benefits of reduced nitrogen input in minimizing nitrate leaching, while frequent applications effectively support corn growth. Furthermore, a common trend emerges when we align nitrogen applications with weather data, particularly focusing on precipitation: Both policies avoid fertilizing on rainy days to prevent nitrate leaching effectively.

In accordance with a 1999 report titled 'Iowa Crop Performance Test- Corn District 2,' published by Iowa State University, corn yielded for various brands ranged between 146 bu/acre and 192 bu/acre, with an average yield of 169.4 bu/acre. It is worth noting that these numbers represented yields of wet corn. To make accurate comparisons with our simulation results, these wet yields must be converted to dry corn yields. The report specified an average moisture content of 17.5\% for the wet corn. In our conversion process, we use a moisture content of 14\% for dry corn, as referenced in a study by Yakoub \textit{et al.} \cite{Yakoub2017}. After converting to dry yield, the range of corn yield in 1999 spans from 8507.53 kg/ha to 11197.74 kg/ha, with an average yield of 9868.59 kg/ha. 

Our simulation results, averaging approximately 9243 kg/ha from optimal policies in Table~\ref{table:1999outcomes}, exhibit a deviation of about 6.4\% from the actual yield. This aligns with variations observed in some prior studies (e.g., 13\% and 9.5\%) that compared the DSSAT simulation results with the actual crop productions \cite{Yakoub2017, Nouna2000}. It is worth noting that the DSSAT maize models, specifically CERES, were initially published 37 years ago \cite{Jones1986}. As such, they may not perfectly capture modern maize production methods, which have evolved substantially over the years. Nonetheless, the optimal policies derived from POMDP models recommend reduced nitrogen usage (as shown in Table~\ref{table:1999outcomes}) compared to the second expert policy, which we will use as a baseline for our subsequent studies in this paper.

In order to assess the influence of nitrate leaching on both nitrogen fertilization practices and corn yield, we conduct a series of simulations to obtain optimal policies from the POMDP-10 model involving different values of $w_3$, ranging from 0 to 50 times the value of $w_2$. Throughout the comparisons, the overall quantity of nitrogen usage remains constant, while the timing of fertilization applications is adjusted according to the specified $w_3$ values. Upon closer examination, we find that assigning a higher weight to nitrate leaching in the reward function (as defined in Equation~(\ref{eq:AgReward})) leads to a reduction in nitrate leaching. This reduction is achieved by minimizing fertilization on rainy days whenever possible. For the subsequent studies detailed in this paper, we choose to maintain $w_3$ five times of $w_2$ in the reward function. This decision aligns with the approach used in a previously referenced study \cite{Wu2022}.

\section{Impact of climate variability on agriculture management}

In this section, we utilize weather data from 1999 as a baseline. Subsequently, we introduce variations in temperature and precipitation to analyze the influence of climate variability. In addition, we delve into the agricultural ramifications of specific extreme events, notably the 1983 heatwave and the 1988 drought. We also examine how optimal policies of fertilization management adapt in response to these events. It is important to note that the soil data used in all simulations remains consistent with the conditions observed in 1999.

\subsection{Impact of higher temperature}

Over the past 70 years, the global climate patterns have undergone significant changes primarily driven by anthropogenic activities, with temperature increases being one of the most prominent aspects \cite{Rabatel2013}. Historical data reveal a steady rise in average temperatures, with a 0.98-degree Celsius increase since 1880. Notably, this warming trend has accelerated in recent decades, with a 0.94-degree Celsius rise recorded in the last 60 years alone. In this study, we systematically elevated the daily average temperature from the 1999 baseline by increments of 0.5, 1, 2, and 5 degrees Celsius \cite{NASA} throughout the year while following a consistent pattern. The precipitation remains the same as in 1999, so we can investigate the impact of temperature variation on fertilization management and corn yield.

In this section, we examine two categories of optimal policies. The first category is the `1999 policy,' which replicates the optimal policy learned by the agent in the POMDP-10 model under the actual weather data from 1999, as outlined in Table~\ref{table:1999outcomes}. The `1999 policy' remains unchanged even when subjected to `hotter' weather conditions, allowing us to assess its adaptability to elevated temperatures. It's important to note that this policy will generate different fertilizer management plans under varying weather conditions. The second category of policy consists of `optimal policies' that the agent re-learns in response to elevated temperatures. In addition, we will investigate Expert policy 2 as detailed in Table~\ref{table:1999outcomes}.

Table~\ref{table:TempImpact} presents the agricultural outcomes, including corn yield, nitrogen input, and nitrate leaching, based on different fertilization management policies. The table illustrates that a 0.5-degree Celsius increase in temperature corresponds to a boost in corn yield and a slight uptick in fertilizer consumption (by the `optimal policy'). As temperatures continue to rise, corn yields also increase, while fertilizer usage begins to drop slightly. This trend is likely due to the temperature approaching the optimal growth range for corn as it escalates.

\begin{table}
\centering
\begin{tabular}{||c c c c||} 
 \hline
 temperature increment & 1999 policy & Optimal policies & Expert policy\\ [0.5ex] 
 \hline
 \hline
 $+0^{\circ}$C &   &    &  
 \\ 
 \hline
 Yield (kg/ha) & 9243 &  9243 & 9247
 \\
  \hline
 Nitrogen input (kg/ha) & 180 &  180 & 224
 \\
  \hline
 Nitrate leaching (kg/ha) & 0.12 &  0.12 & 0.26
 \\
  \hline  \hline
  $+0.5^{\circ}$C &   &    &  
 \\ 
 \hline
 Yield (kg/ha) & 9784 & 10295   &  10303
 \\
  \hline
 Nitrogen input (kg/ha) & 180 & 190   & 224
 \\
  \hline
 Nitrate leaching (kg/ha) & 0.11 & 0.10   &  0.18
 \\
  \hline  \hline
  $+1^{\circ}$C &   &    &  
 \\ 
 \hline
 Yield (kg/ha) & 10416 & 10425   &  10426
 \\
  \hline
 Nitrogen input (kg/ha) & 170 & 160   & 224
 \\
  \hline
 Nitrate leaching (kg/ha) & 0.10 & 0.10   &  0.12
 \\
  \hline  \hline
   $+2^{\circ}$C &   &    &  
 \\ 
 \hline
 Yield (kg/ha) & 9352 & 9357   &  9337
 \\
  \hline
 Nitrogen input (kg/ha) & 140 & 120   & 224
 \\
  \hline
 Nitrate leaching (kg/ha) & 0.09 & 0.09   &  0.09
 \\
  \hline  \hline
  $+5^{\circ}$C &   &    &  
 \\ 
 \hline
 Yield (kg/ha) & 4901 & 4873   &  4901
 \\
  \hline
 Nitrogen input (kg/ha) & 250 & 60   & 224
 \\
  \hline
 Nitrate leaching (kg/ha) & 0.08 & 0.08   &  0.08
 \\
  \hline  \hline

\end{tabular}

\caption{Comparison of different policies when temperature increases.}\label{table:TempImpact}
\end{table}

Our simulations run from April 10th to October 30th, with an average air temperature of 10.9 degrees Celsius in 1999 on the planting day (May 27th), aligning with the recommended corn planting temperature of above 10 degrees Celsius by experts \cite{Abendroth2017}. However, when temperatures surge by more than 2.5 degrees, corn yield starts to decline, possibly indicating that the heat becomes too intense for healthy corn growth. This hypothesis gains further support when the temperature is raised by 5 degrees Celsius, resulting in a significant drop in corn yield to just half its original volume under the actual weather conditions of 1999.

While examining the data presented in Table~\ref{table:TempImpact}, it becomes evident that various policies exhibit similar trends in corn yield as daily temperature systematically increases. However, notable differences emerge in nitrogen input when corn yield starts to decline significantly due to substantial temperature escalation. The `optimal policies,' which consistently employ significantly less nitrogen than the `1999 policy,' lead to higher rewards. Specifically, when the temperature increases by 5 degrees, the 'optimal policy' yields a reward (representing the net income) 30\% higher than the `1999 policy.'

These results underscore the adaptability of the fixed policy to small temperature escalation, while also highlighting the need of the agent to update the optimal policy under extreme temperature conditions. This finding has important implications for PA in response to climate change in the future. It suggests that as temperatures continue to rise, there will be a growing need to develop and implement more dynamic and responsive agricultural management policies to maximize yields and minimize environmental impacts.

\subsection{Impact of insufficient precipitation}

We also examine the influence of rainfall on corn yield and fertilization management. After analyzing the historical rainfall data since 1950, we find no consistent linear pattern in yearly rainfall. For this study, we still use the actual weather conditions of 1999 as a reference and reduce daily rainfall by 20\%, 35\%, 50\%, 65\%, and 80\% throughout the year while maintaining a consistent pattern. Temperatures remain the same as in 1999. Notably, we choose not to increase precipitation, as doing so might introduce the risk of flood damage to the crops, which falls beyond the predictive capabilities of DSSAT.

\begin{table}
\centering
\begin{tabular}{||c c c c||} 
 \hline
 Precipitation reduction & 1999 policy & Optimal policies & Expert policy\\ [0.5ex] 
 \hline
 \hline
 0\% &   &    &  
 \\ 
 \hline
 Yield (kg/ha) & 9243 &  9243 & 9247
 \\
  \hline
 Nitrogen input (kg/ha) & 180 &  180 & 224
 \\
  \hline
 Nitrate leaching (kg/ha) & 0.12 &  0.12 & 0.26
 \\
  \hline  \hline
   20\% &   &    &  
 \\ 
 \hline
 Yield (kg/ha) & 8652 &  8930   &   9108
 \\
  \hline
 Nitrogen input (kg/ha) & 180 &  160   & 224
 \\
  \hline
 Nitrate leaching (kg/ha) & 0.006 &  0.008   &   0.115
 \\
  \hline  \hline
  35\% &   &    &  
 \\ 
 \hline
 Yield (kg/ha) & 8192 &  8408   &   8808
 \\
  \hline
 Nitrogen input (kg/ha) & 180 &  160  & 224
 \\
  \hline
 Nitrate leaching (kg/ha) & 0.006 & 0.008    &   0.009
 \\
  \hline  \hline
   50\% &   &    &  
 \\ 
 \hline
 Yield (kg/ha) & 7164 &  7350   &   7604
 \\
  \hline
 Nitrogen input (kg/ha) & 180 &  130   & 224
 \\
  \hline
 Nitrate leaching (kg/ha) & 0.001 &   0.001  &   0.009
 \\
  \hline  \hline
  65\% &   &    &  
 \\ 
 \hline
 Yield (kg/ha) & 4756 &  5658   &   5587
 \\
  \hline
 Nitrogen input (kg/ha) & 180 & 120  & 224
 \\
  \hline
 Nitrate leaching (kg/ha) & 0.001 & 0.001    &   0.001
 \\
  \hline  \hline
  80\% &   &    &  
 \\ 
 \hline
 Yield (kg/ha) & 2406 &  4360   &   4025
 \\
  \hline
 Nitrogen input (kg/ha) & 180 &  100   & 224
 \\
  \hline
 Nitrate leaching (kg/ha) & 0.001 & 0.001   &   0.001
 \\

  \hline  \hline

\end{tabular}

\caption{Comparison of different policies when precipitation decreases.}\label{table:RainImpact}
\end{table}

Table~\ref{table:RainImpact} presents the simulated outcomes under varying precipitation scenarios. Similar to Table~\ref{table:TempImpact}, this table compares corn yield, nitrogen input, and nitrate leaching between various policies, including the `1999 policy,' optimal policies, and an expert policy.  It can be seen that when precipitation decreases, the corn yield is significantly impacted. The corn yield can be less than half the standard value when the weather is severely dry. Given that corn is a moisture-intensive crop \cite{Paolo2008}, the results are convincing. 

In addition, while adhering to the `1999 policy,' fertilization practices remain unchanged. However, the `optimal policies' utilize less fertilizer due to reduced precipitation, which doesn't significantly impact nitrate leaching. We also include the results from the expert policy. In line with the earlier simulation results regarding temperature variability, both the optimal policies and the expert policy yield similar corn productions. The primary distinction between the two policies is the quantity of nitrogen applied.

Overall, Table~\ref{table:RainImpact} also illustrates that optimal policies consistently outperform the `1999 policy' in corn yield, especially in conditions with significantly low rainfall. This finding aligns with our conclusion in the study of the impact of higher temperatures on agriculture and agricultural management. The optimal policy learned under typical weather conditions demonstrates adaptability when faced with minor precipitation fluctuations. However, in the case of a significant reduction in precipitation, the agent must acquire a new optimal policy. When comparing Table~\ref{table:RainImpact} with Table~\ref{table:TempImpact}, it becomes evident that reductions in precipitation have a more pronounced impact than temperature increases. This highlights the crucial role that humidity levels play in corn cultivation. It is important to note that this study doesn't consider irrigation as a factor.

The notably low nitrate leaching values of 0.001 kg/ha observed under both severe precipitation reductions may reflect not only the physical constraints of water-limited nitrogen movement in soils but also a limitation of the DSSAT simulator under such extreme environmental changes. DSSAT’s leaching algorithms are primarily parameterized for conditions within a plausible range of soil moisture and temperature dynamics derived from historical and experimental datasets. When climate conditions deviate far beyond these calibration bounds the model may underestimate nitrate transport because simulated soil water fluxes become effectively negligible. In reality, even under such stressed conditions, small but measurable nitrate losses can occur through occasional rainfall events, preferential flow in cracked soils, or subsurface lateral movement. These pathways may not be fully captured by DSSAT’s standard routines without site-specific calibration. For this reason, we retain the model’s output of 0.001 kg/ha as the reported value, acknowledging it as the simulator’s minimal positive output rather than a definitive measurement of field leaching.

\subsection{Heat wave and drought}

To further study the impact of extreme weather events on agriculture and fertilization management, we consider two real scenarios that occurred in Iowa: the heat wave in 1983 and the drought in 1988. Previous research \cite{Motha2005} indicated that these extreme weather events led to 32\% and 38.5\% reduction in Iowa's corn yields compared to the previous years. 

To accurately simulate these events, we source data from the Iowa Environmental Mesonet (IEM) for daily maximum and minimum temperatures, as well as precipitations, for the years 1982, 1983,  1987, and 1988. Additionally, we gather information on planting and harvesting dates from the 'THE 1983 IOWA CORN YIELD TEST REPORT District 2' and  'Iowa Corn Yield Test Report (Iowa State University) District 1 December 1988.' According to the reports, corn was planted on May 7th \& 8th and harvested on October 20th \& 21st in 1983, and on May 3rd and harvested on October 4th \& 5th in 1988.

\begin{figure}
\centering
\resizebox*{8cm}{!}{\includegraphics{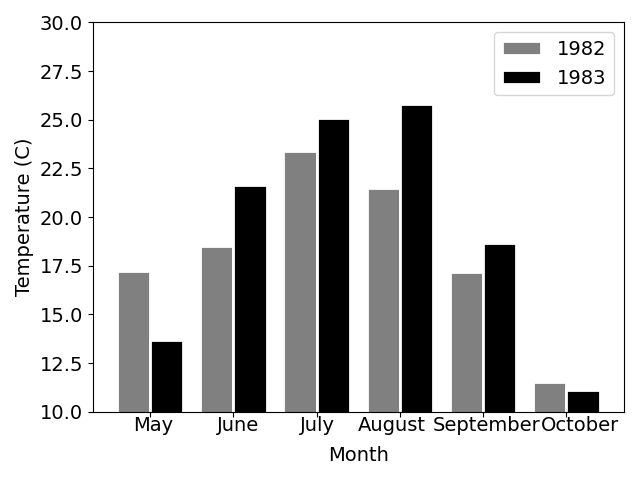}}
\caption{Monthly average temperatures from May to October in 1982 and 1983.} \label{fig:8283t}
\end{figure}

\begin{figure}
\centering
\resizebox*{8cm}{!}{\includegraphics{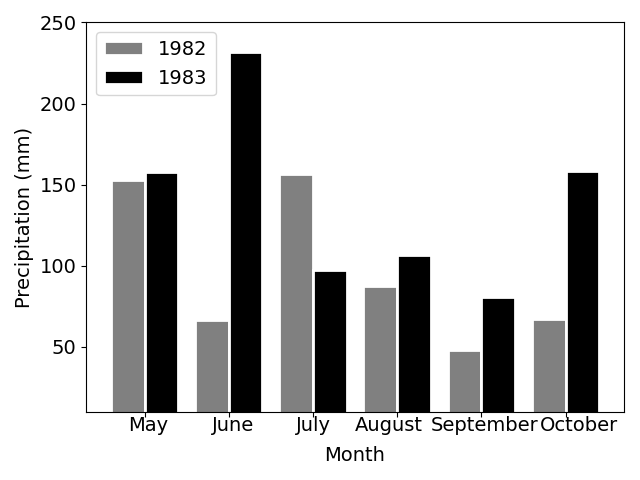}}
\caption{Monthly total precipitations from May to October in 1982 and 1983.}  
\label{fig:8283p}
\end{figure}

Figures ~\ref{fig:8283t} and~\ref{fig:8283p} depict the comparisons of monthly average temperatures and precipitations for the months of April through October in 1982 and 1983. It can be seen that the average temperatures in June, July, August, and September of 1983 were higher than in 1982 by 3.1 degrees, 1.7 degrees, 4.3 degrees, and 1.5 degrees, respectively. However, the overall precipitation in 1983 was higher than in 1982, especially in June 1983. Therefore, there was a heatwave in 1983 but not a drought.

\begin{figure}
\centering
\resizebox*{8cm}{!}{\includegraphics{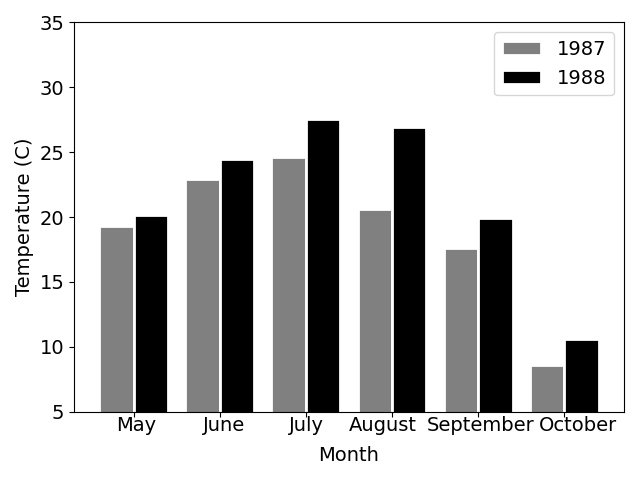}}
\caption{Monthly average temperatures from May to October in 1987 and 1988.} 
\label{fig:8788t}
\end{figure}

\begin{figure}
\centering
\resizebox*{8cm}{!}{\includegraphics{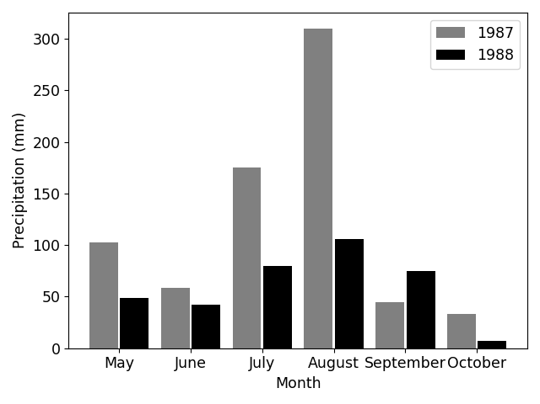}}
\caption{Monthly total precipitation from May to October in 1987 and 1988.} 
\label{fig:8788p}
\end{figure}

Figure~\ref{fig:8788t} and Figure~\ref{fig:8788p} compare monthly average temperatures and precipitation between 1987 and 1988. The average temperature in August of 1988 was significantly higher than in 1987, with a difference of 6.3 degrees, larger than those in August of 1982 and 1983. However, the precipitation in August was considerably lower in 1988 compared to 1987, representing approximately a 400\% reduction in rainfall. Although there were more rains in July and September of 1988 compared to 1987, the severe reduction in precipitation during August, a crucial corn growth stage, indicates that Iowa experienced a drought in 1988, emphasizing the significance of this dry period more than the heatwave. 

Our simulations utilize the actual weather data of 1982, 1983, 1987, and 1988 while maintaining the soil data consistent with 1999. To compare corn yield and nitrogen input between 1982 and 1983, the agent first learns an optimal policy for 1982, which is then applied to 1983. This is compared to the optimal policy that the agent learns specifically for 1983. We follow the same procedure for the comparison between 1987 and 1988. The results of these comparisons are presented in Table~\ref{table8288}.

\begin{table}
\centering
\begin{tabular}{||c c c c ||} 
 \hline
Year& Policy & Corn yield (kg/ha) & Nitrogen input (kg/ha) \\ [0.5ex] 
 \hline\hline
 \hline
 1982 & Optimal policy  & 10923 & 270
 \\ 
  \hline
 1983  & 1982 policy & 7318 & 320
 \\
 \hline
 1983 & Optimal policy & 8098 & 110
 \\
 \hline 
  \hline\hline
 1987 & Optimal policy &9963  &180
 \\ 
  \hline
 1988  & 1987 policy & 3820  & 200
\\
\hline
 1988 & Optimal policy  & 4344 & 130
 \\
 \hline 
\end{tabular}

\caption{Comparison of simulation results for 1982 and 1983 \& 1987 and 1988}\label{table8288}
\end{table}

An optimal policy of fertilization management is learned for 1982, and its effects on corn yield and nitrogen input are documented in Table~\ref{table8288}. However, when the same policy is applied in 1983, we observed a drop of 33\% in production due to the heatwave. This decline closely mirrors historical data from the USDA for Iowa, which indicates a 32\% decrease in corn production in 1983 compared to the previous year (1982). This suggests that the fertilization strategy developed in 1982 may not have been well-suited to handle the extreme weather event, i.e., heatwave, in 1983.

Interestingly, despite a significant decrease in corn yield, nitrogen input increases by 18.5\% in 1983 compared to 1982 when using the `1982 policy.' This suggests that the fertilization application is not adjusted to match the unique conditions of 1983, potentially leading to an excessive use of nitrogen. The agent learns a separate optimal policy specifically for 1983 to address these challenges. The induced fertilization strategy results in a higher corn yield by 10\% while significantly reducing nitrogen input. 

A similar pattern is observed when assessing the impact of the 1988 drought on agricultural outcomes. When the optimal policy learned for 1987 is applied in 1988, it results in a significant reduction in corn production, amounting to a 61.5\% decrease, as depicted in Table~\ref{table8288}. However, when the optimal policy specifically tailored by the agent for 1988 is employed, the production reduction was slightly less severe at 56\%, but it came with a substantial reduction in nitrogen input. 

It is important to note that the simulated reduction in corn yield, as seen in the study, greatly exceeded the actual drop, which is 38.5\%. This discrepancy can be attributed to the fact that this study doesn't account for irrigation in the agricultural management strategy. Corn has a high water requirement and is particularly sensitive to drought, so the omission of irrigation likely contributed to the larger simulated yield reduction.

\section{Conclusion and future works }

Optimizing crop management strategies is essential for maximizing yield, reducing costs, and mitigating environmental impacts. In this study, we introduced a framework that combines DRL with RNNs, utilizing Gym-DSSAT to determine optimal nitrogen fertilization strategies. Our findings reveal that the agricultural environment, as represented by Gym-DSSAT, is partially observable. This differs from the assumptions made in previous studies, where the state of the agricultural environment was assumed to be entirely determined from the currently observed internal variables provided by Gym-DSSAT. To address this challenge, we compared POMDP models to MDP models. Our results indicated that leveraging a sequence of observations allows the agent to learn and implement more effective policies for nitrogen fertilization management. 

We also applied our developed framework to assess the impacts of climate variability on agricultural outcomes and management, with a particular focus on scenarios involving higher temperatures and inadequate precipitation. We found that the pre-learned optimal policy proves adaptable under minor climate variability but falls short in extreme weather conditions. This study underscores the critical importance of tailoring fertilization management practices to the specific weather conditions of each year, especially in the face of extreme weather events. Such adaptability is essential for optimizing crop yield while simultaneously minimizing nitrogen input. It recognizes the need for agriculture to maintain flexibility in response to the variable and, at times, extreme influences of weather and other factors on crop performance. 

Due to data limitations, the simulations presented in this paper rely solely on the 1999 dataset, particularly for soil properties. Going forward, we aim to compile a comprehensive historical soil dataset for the relevant agricultural lands. With this enriched dataset, we aspire to conduct more representative and accurate simulations, thereby enhancing our findings' precision and reliability. 

While this paper primarily focused on nitrogen fertilization, our future research will incorporate irrigation management, particularly in addressing severe drought events. Furthermore, we intend to gather comprehensive cost data for the relevant year, encompassing machine, labor, and other expenses. Integrating these variables into our reward function will enable our model to replicate farmers' net incomes more accurately.

When studying the impact of temperature and precipitation variability on agriculture, we maintain the same patterns consistent with those observed in 1999. We acknowledge the limitation of solely relying on historical data for simulations. In subsequent research, we plan to generate random weather scenarios based on real data to introduce weather uncertainty and perform uncertainty quantification of agricultural management and outcomes. 

\section*{Author contributions statement}
Zhaoan Wang. Shaoping Xiao. and Jun Wang. Conceptualization, 
Zhaoan Wang. Data curation, 
Zhaoan Wang. Investigation, 
Zhaoan Wang. Shaoping Xiao. and Junchao Li. Methodology, 
Zhaoan Wang. Writing – original draft,
Shaoping Xiao. Supervision, 
Shaoping Xiao. and Jun Wang. Funding acquisition, 
Shaoping Xiao. Junchao Li. and Jun Wang. Writing - review \& editing. 

All authors reviewed the manuscript.

\section*{Declaration of Competing Interest}
The authors declare that they have no known competing financial interests or personal relationships that could have appeared to influence the work reported in this paper.

\section*{Data availability}
The data used in this study are freely available. Some data files and codes are available on the GitHub site 
(https://github.com /ZhaoanWang /Learning-based-agricultural-Management). You can contact the corresponding author, Mr. Zhaoan Wang (zhaoan-wang@uiowa.edu) if you want to use the data. 

\section*{Funding}
This material is based upon work supported by the U.S. Department of Education under Grant Number ED\#P116S210005 and the National Science Foundation under Grant Numbers 2226936 and 2420405. Any opinions, findings, and conclusions or recommendations expressed in this material are those of the author(s) and do not necessarily reflect the views of the U.S. Department of Education and the National Science Foundation. 

\section*{Acknowledgments}
The authors acknowledge the support from the University of Iowa OVPR Interdisciplinary Scholars Program for this study.

\section*{Declaration of Generative AI and AI-assisted technologies in the writing process}
During the preparation of this work, the authors used GPT-3.5 in order to improve the readability and language during the writing process. After using this tool/service, the authors reviewed and edited the content as needed and took full responsibility for the content of the publication.

\end{document}